\documentclass{article}

\usepackage{arxiv}
\usepackage[numbers]{natbib}

\usepackage{amsmath}
\usepackage{enumitem}

\usepackage[utf8]{inputenc} % allow utf-8 input
\usepackage[T1]{fontenc} % use 8-bit T1 fonts
\usepackage{hyperref} % hyperlinks
\usepackage{url} % simple URL typesetting
\usepackage{booktabs} % professional-quality tables
\usepackage{amsfonts} % blackboard math symbols
\usepackage{microtype} % microtypography
\usepackage{lipsum}
\usepackage{fancyhdr} % header
\usepackage{graphicx} % graphics
\graphicspath{{media/}} % organize your images and other figures under media/ folder

\DeclareMathOperator{\var}{var}
\DeclareMathOperator{\cov}{cov}

\usepackage{xcolor}

\newtheorem{lemma}{Lemma}

\title{Challenges for Predictive Modeling with Neural Network Techniques using Error-Prone Dietary Intake Data.\protect}

\author{
 Dylan Spicker \\
  Department of Mathematics and Statistics\\
  University of New Brunswick (Saint John)\\
  New Brunswick, Canada \\
  \texttt{dylan.spicker@unb.ca} \\
   \And
 Amir Nazemi \\
  Department of Systems Design Engineering\\
  University of Waterloo\\
  Ontario, Canada \\
  \And
 Joy Hutchinson \\
  School of Public Health and Health Systems\\
  University of Waterloo\\
  Ontario, Canada \\
  \And
Paul Fieguth \\
  Department of Systems Design Engineering\\
  University of Waterloo\\
  Ontario, Canada \\
   \And
Sharon I. Kirkpatrick\\
  School of Public Health and Health Systems\\
  University of Waterloo\\
  Ontario, Canada \\
  \And
Michael Wallace \\
  Department of Statistics and Actuarial Science \\
  School of Public Health and Health Systems\\
  University of Waterloo\\
  Ontario, Canada \\
   \And
 Kevin W. Dodd \\
 Biometry Research Group, Division of Cancer Prevention \\
 National Cancer Institute \\
 Maryland, USA
}

\begin{document}
% \presentaddress{This is sample for present address text this is sample for present address text}

\maketitle

\begin{abstract}
Dietary intake data are routinely drawn upon to explore diet-health relationships and inform clinical practice and public health. However, these data are almost always subject to measurement error, distorting the true diet-health relationships. Beyond the measurement error, there are likely complex synergistic and sometimes antagonistic interactions between different dietary components, complicating the relationships between diet and health outcomes. Flexible models are required to capture the nuance that these complex interactions introduce. This complexity makes research on diet-health relationships an appealing candidate for the application of modern machine learning techniques, and in particular, neural networks. Neural networks are computational models that are able to capture highly complex, nonlinear relationships so long as sufficient data are available. While these models have been applied in many domains, the impacts of measurement error on the performance of predictive modeling has not been systematically investigated. However, dietary intake data are typically collected using self-report methods and are prone to large amounts of measurement error. In this work, we demonstrate the ways in which measurement error erodes the performance of neural networks, and illustrate the care that is required for leveraging these models in the presence of error. We demonstrate the role that sample size and replicate measurements play on model performance, indicate a motivation for the investigation of transformations to additivity, and illustrate the caution required to prevent model overfitting. While the past performance of neural networks across various domains make them an attractive candidate for examining diet-health relationships, our work demonstrates that substantial care and further methodological development are both required to observe increased predictive performance when applying these techniques, compared to more traditional statistical procedures.
\end{abstract}

\keywords{measurement error, prediction, artificial neural networks, dietary data, machine learning, dietary assessment}

\footnotetext{\textbf{Abbreviations:} MSE, mean squared error; MSPE, mean squared prediction error; SGD, stochastic gradient descent}

\section{Introduction}
Measurement error describes scenarios where the observed value of a variate differs from the value we wish to observed. While measurement error is pervasive across many disciplines, it is of particular concern in the analysis of dietary intake data, where reported dietary intake will often differ from true dietary intake \cite{Thompson2015,Keogh_2020}. Dietary intake data are often collected from one of two instruments: food-frequency questionnaires (FFQs) or 24-hour recalls (24HRs). FFQs are questionnaires which solicit responses regarding the broad frequency with which particular foods are consumed by individuals, with typical recall periods ranging from a month to a year \cite{Cade2002,Thompson2015}. In contrast, 24HRs ask individuals to recall what foods were consumed during a specific 24 hour period \cite{baranowski201224}. The values measured from instruments directly are referred to as \emph{reported intake}, and both 24HRs and FFQs produce reported intake data that is subject to measurement error \cite{Kirkpatrick_2022}. Typically, 24HRs have less systematic error than FFQs, however, a 24HR merely captures a snapshot in time \cite{freedman2014pooled,freedman2015pooled}. As a result, repeated measurements of 24HRs are often taken in order to estimate \emph{usual intake} (i.e., the long-term average intake) of a dietary component. When considering the impact of dietary intake on health outcomes, we most often want to understand the impact of usual intake rather than reported intake. The effects of measurement error on estimation and inference have been thoroughly studied, and the impacts of ignoring measurement error are well understood for many commonly applied statistical estimation techniques \cite{Keogh_2020}.

In many settings, the primary focus of an analysis is on prediction rather than estimation directly. For instance, it is of interest to develop predictive models for the development of breast cancer \cite{BreastCancer}, to be able to preemptively detect seizures in patients with epilepsy \cite{Seizures}, to use dietary information to predict coronary heart disease \cite{CHD_CNN}, or to predict verbal intelligence during adulthood from adolescent physical activity and nutrition \cite{AdolescentNutrition}. In these analyses, the primary aim is developing a model with predictive utility rather than explanatory power. As a result, the common concerns of consistency and bias are subordinated to the predictive performance of the models.

The measurement error literature tends to focus on reducing bias or restoring consistency in estimation, and on correcting uncertainty quantification for inference, with comparatively less attention paid to problems of prediction. Part of the reason for this is well summarized in an argument put forward by Carroll et al. (Section 2.6)\cite{MeasurementErrorBook1}. The authors suggest measurement error is of limited concern in prediction problems, since the prediction problem can be reframed to use the error-prone versions as predictors directly. Consequently, it is statistically convenient to model the outcome of interest from the observed quantities, even if those happen to be error-prone. In the context of dietary intake, for instance, this may mean predicting an individual's systolic blood pressure based not on their usual sodium intake, but rather their reported sodium intake.

For the use of observed quantities to be an effective strategy the analyst must be able to specify a useful predictive model linking the outcome and the error-prone, surrogate measurements. This can be non-trivial since it will generally be the case that a naive substitution of the surrogate measurements into the standard outcome model will result in poor performance. For instance, if a linear regression is thought to be suitable to predict the outcome from the true, error-free variate, there is no guarantee that a linear regression is appropriate when using the surrogate measurement. There may also be substantial loss of predictive performance when building models using error-prone surrogates compared to the true, error-free variates. In the context of risk prediction, Khudyakov et al.\cite{RiskPrediction} demonstrate that the use of error-prone variates in place of the true variate can reduce the performance of predictive models substantially. This reduction in performance may be of particular concern if these limitations are not properly understood, and the authors recommend consideration when designing studies where there is the possibility of taking more costly measurements that are subject to less error, on fewer individuals. 

In dietary assessment research, depending on what dietary components are of primary interest, it is often the case that there is no practical way to obtain even an asymptotically error-free measure of true intake. One technique is to directly observe individuals over time, however, this is not feasible over long periods of time or with large samples. Beyond observation, there are four available ``recovery biomarkers'' - assessment instruments subject exclusively to random noise, and thus able to use averaging to approximate true intake. These are: doubly labelled water to assess total energy intake, 24-hour urinary sodium excretion to assess sodium intake, 24-hour urinary potassium excretion to assess potassium intake, and 24-hour urinary nitrogen excretion to assess protein intake \cite{Freedman2011}. These biomarkers assess total intake of the specific components they measure, however, they cannot assess the total diet and are both burdensome and expensive to administer. Thus, researchers generally rely on self-report assessment instruments that can capture total dietary intake but are prone to both bias and noise. When using 24HRs, researchers may take several repeated measurements across different days to better estimate usual intake. While this procedure can partially ameliorate concerns relating the error and bias in usual intake estimation, increasing the number of observations per individual results in high participant burden, and the quality of responses decreases as the number of responses increases \cite{Dodd_2006}. Of final note, in order for the argument regarding ignoring errors in predictive modeling to hold, it must be the case that the measurement error models are \emph{transportable}. If the data from one population are used to construct a predictive model that is to be applied in a different population, then the modeling strategy based on surrogate measurements is only valid when the errors in each population are drawn from the same distribution. 

While the loss of predictive performance and the requirement of transportability are innate features of the statistical structures under consideration, the need for correct model specification can be addressed through the careful selection of modeling strategies. An analyst may, for instance, choose nonparametric techniques that can capture the underlying relationship effectively, regardless of the true complexity of that relationship. In light of modern advances in computer science, one strategy that has demonstrated particular efficacy in complex modeling scenarios is the use of \emph{artificial neural networks}. Artificial neural networks (hereafter referred to as simply neural networks) are computational models, loosely designed to mimic the structure of neurons in a brain, which are commonly deployed for the purpose of machine learning \cite{lecun2015deep,abiodun2018state}. From a statistical perspective, neural networks can be viewed as either a semiparametric or nonparametric technique, where sufficient parameters can be introduced to capture highly complex relationships, without the need for direct specification. As a result of advances in the computational feasibility of \emph{deep neural networks} (that is, neural networks with very large numbers of parameters), and with growing access to data, neural networks have been successfully employed in many complex prediction settings. For instance, neural networks have been used to predict crop yields from predicted weather data \cite{Khaki_2019}, reaction types in organic chemistry \cite{Wei_2016}, user responses in online advertising contexts \cite{Qu_2016}, diet quality through a healthy eating index score \cite{Hearty2008AnalysisOM}, and clinical events using electronic health record data \cite{PMLR}. Neural networks are also able to incorporate diverse types of data, both as explanatory variates and outcomes, and are able to make effective use of very large data. The promise of using neural networks, and machine learning more broadly, to allow for flexible, complex, predictive models has generated interest in the context of modeling diet-health relationships. 

Neural networks are effectively a ``black box algorithm'', in that it is difficult to credibly understand the rationale for generating a particular prediction \cite{Montavon2018}. As a result, it is critical to validate the specific underlying assumptions prior to the application of neural networks in any decision-making context. In this work, we provide the first quantitative consideration of the impacts of measurement error on neural network predictive models, informed by considerations relevant to dietary intake data. We demonstrate, through extensive simulations, possible concerns with the application of  machine learning techniques in this context without careful consideration of error mechanisms. Our results suggest that, while there is substantial promise in the use of neural networks to capture the complex relationships within diet and health relationships, the same types of concerns that arise in traditional statistical modeling of these relationships are present when using neural networks, and need to be carefully considered. 

Specifically, we demonstrate how transformations of error-prone data can, in certain settings, improve predictive performance in dietary models, we illustrate the trade-offs between an increase in sample size and an increase in the number of measurements taken for each individual, and we demonstrate concerns relating to model overfitting which are accentuated in the presence of measurement error. Taken together, our results demonstrate that, without careful consideration of the underlying data structure, advanced computational techniques do not provide a notable benefit in predictive performance, and come at the cost of lessened interpretability and increased computational time. The results also suggest that, with care, there is good reason to pursue the use of machine learning techniques for the analysis of dietary intake data. Our results indicate the need for care in the application and deployment of neural networks to predict diet-health relationships, and provide the foundation for future developments in the application of machine learning to understanding such relationships. 

This article proceeds by introducing measurement error and neural networks, generally, before discussing how the former interacts with the latter in the context of dietary data, in Section~\ref{sec::methodologies}. We then illustrate, heuristically and theoretically, the concerns relating to using neural networks with error prone data in Section~\ref{sec::theory}. In Section~\ref{sec::numerical_studies}, the simulation experiments investigating the performance of neural networks subject to measurement error are detailed, and the results from these studies are provided in detail. The simulations form a case study based on data from the National Health and Nutrition Examination Survey (NHANES), rooting the analyses in real-world scenarios. Finally, Section~\ref{sec::discussion} provides a discussion of the practical implications of our results. 

\section{Methodologies}\label{sec::methodologies}
We start by introducing the notation and measurement error models under consideration, the mathematical foundations of neural networks, and reviewing the use of machine learning in nutritional epidemiology. In Section~\ref{subsec::outcome_mem} we present the notation and underlying assumptions for the framework of error-prone dietary intake data. We assume that outcomes are from a generalized regression framework, where some subset of the predictors are measured with error. Following that, Section~\ref{subsec::predictive_models} introduces the statistical framework of predictive modeling, with a specific focus on the mathematical definition of neural networks. Finally, Section~\ref{subsec::background} explores the existing applications of machine learning techniques in dietary research. 

\subsection{Outcome and Measurement Error Models}\label{subsec::outcome_mem}
Consider a response variable, $Y$, which is related to predictors $\{X, Z\}$. We take $X$ to be the predictors that are error-prone and as such may not be observed accurately, while $Z$ contains all other factors that are measured without error. We will assume that there are $p$ error-prone predictors, and $q$ error-free predictors. In the simulations that follow, we will take a generalized regression framework, where we assume that \begin{equation}
 E[Y|X, Z] = g^{-1}\left(\alpha_0 + \alpha_X'X + \alpha_Z'Z\right), \label{eq::outcome_model}
\end{equation} where $g(\cdot)$ is a known link function, and then $\alpha = (\alpha_0, \alpha_X', \alpha_Z')'$ is a vector of $p+q+1$ unknown parameters, with $\alpha_X \in \mathbb{R}^p$ and $\alpha_Z \in \mathbb{R}^q$. The conditional variance of $Y$ given $\{X,Z\}$ is denoted $\sigma_Y^2$.

In the measurement error literature, in place of $X$ we observe a surrogate measurement denoted $X^*$. We will often assume a specific functional form relating $X$ and $X^*$, with the most common assumption being that $X^* = X + V$, for a random error term $V$, with mean zero, conditional on $X$, and constant variance. This is the \emph{classical additive} measurement error model. In estimation and inference problems, corrections for the effects of measurement error typically rely on auxiliary information, often in the form of \emph{replicate} measurements. Instead of observing a single surrogate measurement, $X^*$, for each individual, several measurements are taken. This gives $\{X^*_1,\dots,X^*_m\}$, which are typically assumed to be independent and identically distributed. Using these repeated measurements, information regarding the size and distribution of the errors can be determined.

In the case of dietary intake data, we take $X$ to represent the usual intake of a particular dietary component, which is rarely directly measurable. Instead, we conceive of the $\ell$-th component of $X$ (denoted $X_{\ell}$) as an \textit{unbiased} measurement of a single day's intake of a particular dietary component, for a particular individual. Then, because true usual intake is the expected value of single day intake, any particular day's reported dietary intake is an error-prone, surrogate measurement of usual intake subject only to noise. Suppose that the reported intake for the $\ell$-th dietary component on day $j$ is denoted $X^*_{\ell,j}$, then for any individual we define the usual intake $X_{\ell} = E[X^*_{\ell,j}|Z]$. Take $f(x)$ to be a Box-Cox transformation\cite{BoxCox}, then we further assume that every recorded daily observation is given by \begin{equation}
 X^*_{\ell,j} = f^{-1}\left(\beta_{0,\ell} + \beta_{Z,\ell}'Z + u_{\ell} + \epsilon_{\ell, j}\right), \label{eq::surrogate_definition}
\end{equation} where $u_{\ell}$ captures the between individual variation for intake of the dietary component $\ell$, and $\epsilon_{\ell,j}$ represents the within person variation with the measurement having been taken on day $j$. $\beta_{\ell} = (\beta_{0,\ell}, \beta_{Z,\ell}')'$ is a $q+1$ dimensional vector of unknown parameters. 

In our numerical experiments, we assume that $U$ is normally distributed with mean zero and variance of $\Sigma_u$, denoting realizations as $u = (u_{1},\cdots,u_{p})'$. Further, for $j=1,\dots,m$, $\epsilon_j = (\epsilon_{1,j},\cdots,\epsilon_{p,j})'$ is assumed to be independent and identically distributed, according to a zero-mean normal distribution with variance $\Sigma_{\epsilon}$.We assume that $U$ and $\epsilon_{\cdot}$ are independent of each other, and of the outcome $Y$. We assume that these observations are made for a sample of size $n$, such that for each $i=1,\dots,n$ the data consist of $\{Y_i,X^*_i,Z_i\}$, where $X^*_i = (X^*_{i,1},\dots,X^*_{i,p})$ and $X^*_{i,j} = (X^*_{i,1,j},\cdots,X^*_{i,p,j})$. In the case of dietary data, the more days' worth of dietary measurements that are taken, the more repeated measurements that are available, and we use these expressions interchangeably. 

When we take $f(\cdot)$ to be the identity, and set $\beta_{Z,\ell} = 0$ then we are are left with $X^*_{i,\ell,j} = \beta_{0,\ell} + u_{i,\ell} + \epsilon_{i,\ell,j}$, where $u_{i,\ell}$ and $\epsilon_{i,\ell,j}$ are both normally distributed and $\beta_{0,\ell}$ is an unknown constant. This is the classical additive model previously described. The added flexibility of dependence on error-free variates and on a transformation allows for more realistic error models to be accommodated in the same underlying framework. 

In this framework, the standard goal of inference and estimation centers on estimating $\alpha$ given the observed data. There are many techniques which exist that allow for the consistent estimation of regression parameters in this setting \cite{Dodd_2006,Tooze_2010}. Instead of interest in $\alpha$ directly, we are concerned with $\alpha$ only insofar as it pertains to the prediction of $Y$. Our analytical goal is to develop a model that takes as input $\{X^*, Z\}$ and outputs a prediction for $Y$. Given this notation, the argument in favor of ignoring measurement error for the purpose of prediction can be expressed succinctly by noting that $X^*$ is an error-free measurement of itself. Then, instead of modeling $Y$ given $\{X,Z\}$, we model $Y$ given $\{X^*, Z\}$. 

The need for flexible modeling in this setting becomes apparent in light of equations~(\ref{eq::outcome_model}) and (\ref{eq::surrogate_definition}). Even supposing that the link functions are known, the interaction of these models is complex, generally meaning that \[E[Y|X^*, Z] \neq g^{-1}(\widetilde{\alpha}_0 + \widetilde{\alpha}_X'\widetilde{h}(X^*) + \widetilde{a}_Z'Z),\] for a vector $\widetilde{\alpha}$ and transformation $\widetilde{h}(X^*)$. An exception to this exists when $g(\cdot)$ is taken to be the identity. This gives a linear model for the conditional expectation of $Y$ given $\{X,Z\}$, and in this case we can write that, \[E[Y|X^*, Z] = E\left[E[Y|X^*, X,Z]|X^*, Z\right] = \alpha_0 + \alpha_X'E[X|X^*, Z] + \alpha_Z'Z.\] This model is linear in \[E[X|X^*, Z] = E[E[X^*_{j}|Z]|X^*, Z] = E[X^*_{j}|Z].\] If $f(\cdot)$ is also the identity, then this will render the entire model linear, which means that $E[Y|X^*, Z]$ takes the same form, up to the parameters, as $E[Y|X,Z]$. 

Even when linear models are assumed, a slight extension of equation~(\ref{eq::outcome_model}) is commonly applied where, instead of taking the outcome to be linear in $X$, some function is applied to the dietary components transforming them. We call this function $h$, such that $h(X)$ returns the relevant predictors of the outcome of interest which are related linearly to $g(E[Y|X,Z])$. This could include, for instance, products or ratios of multiple components, or other similar combinations. In this case we write that \begin{equation}
 E[Y|X, Z] = g^{-1}\left(\alpha_0 + \alpha_{h}'h(X) + \alpha_Z'Z\right), \label{eq::outcome_model_expanded}
\end{equation} where $h\colon\mathbb{R}^{p}\to\mathbb{R}^{k}$ and $\alpha_h$ is a $k\times1$ vector. Using this model, even if $h$ is known and both $g(\cdot)$ and $f(\cdot)$ are linear, it will not generally be the case that $E[Y|X^*, Z]$ is linear in a transformed version of $X^*$.

\subsection{Predictive Models and Neural Networks}\label{subsec::predictive_models}
The goal of the analysis is to construct a function of the observable data that produces estimates of the outcome. That is, we want to construct an estimator, $\widehat{T}(X^*, Z)$ that produces $\widehat{y} = \widehat{T}(X^*, Z)$ as an estimate for $Y$. We are predominantly concerned with the accuracy of predictions generated by $\widehat{T}$ as measured by a loss function of interest, say $\mathcal{L}_{\widehat{T}}(Y, \widehat{y})$. The loss function of interest will depend on the type of outcome that is observed, and the specific context that the prediction model should be used in. In our work, we are predominantly concerned with the mean squared error (MSE). For predictions generated on a set of individuals, $i=1,\dots,n_\text{test}$, the MSE is given by \begin{equation}\mathcal{L}(\widehat{T}) = \mathcal{L}_{\widehat{T}}(Y, \widehat{y}) = \frac{1}{n_\text{test}}\sum_{i=1}^{n_\text{test}}\left[Y_i - \widehat{T}(X^*_i, Z_i)\right]^2.\label{eq::loss_function}\end{equation}

It is worth distinguishing between the loss that is observed within the \emph{training data} and the loss that is observed within the \emph{testing data}. Training data refer to data that are used to estimate $\widehat{T}$, while testing data are data that are withheld from the model fitting procedure used exclusively to test predictive performance. While it is generally preferable to have a small loss on both the training and testing data, in practice, performance on the testing data is of more value. If a model performs substantially better on training data than it does on testing data, we say that the model is \emph{overfitting} the data. Testing data are meant to serve as more representative of the actual use case for the model, where predictions are being generated on previously unseen data. The MSE computed on a validation sample is sometimes referred to as the \emph{mean squared prediction error} (MSPE). 

\begin{figure}
 \centering
 \includegraphics[width=\textwidth]{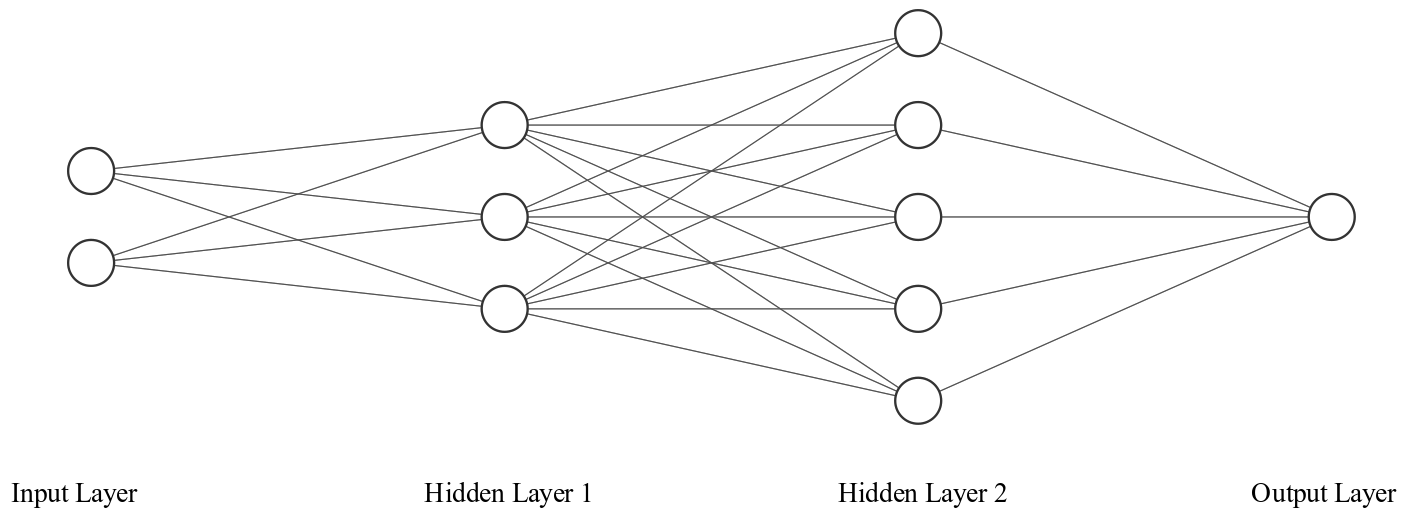}
 \caption{An example architecture for a fully connected neural network. Here the input layer has 2 nodes ($n^{[1]} = 2$), the first hidden layer has 3 nodes ($n^{[2]} = 3$), the second hidden layer has 5 nodes ($n^{[3]} = 5$), and the output layer has a single node ($n^{[4]} = 1$). With this structure, we are using 2 predictors to predict a univariate outcome.}
 \label{fig::nn_structure}
\end{figure}

Any method of model fitting can be applied in this setting. For instance, with a known link function, it may be natural to pursue prediction through generalized linear models or similar regression-based techniques. However, the previously indicated complexity renders techniques that make fewer parametric assumptions particularly amenable to this setting. One such approach is the use of neural networks. Neural networks are well-suited to the task as they are universal approximators, which means that neural networks can approximate any relationship observed in data arbitrarily well, regardless of the complexity of the functional form generating the data \cite{UAT_1,UAT_2}. A neural network is comprised of \emph{nodes} that are arranged into sequential \emph{layers}. The first layer consists of one node for every input to the network. This input layer is connected to one or more \emph{hidden layers}, which are connected in sequence to a final layer which has the same dimension as $Y$ (in the case of a continuous outcome, this is a size of $1$). The hidden layers can contain as many nodes, and there can be as many hidden layers, as is computationally feasible. We denote the size of the $j$-th layer as $n^{[j]}$. An example architecture is presented in Figure~\ref{fig::nn_structure}.

In a fully connected network, each node in a layer is connected via an edge to each node in the next layer. Each of these edges are assigned a particular numeric weight. The value at a particular node is computed by taking the weighted sum of all nodes at the previous layer, weighted by the edge weights connected to the current node, and then applying an \emph{activation function} to the resulting sum. The activation function can be any scalar function, though they are often selected to be nonlinear to introduce additional nonlinearity into the the network. Once the value for a node is computed, it is pushed forward through the network, until the final outcome is produced. 

Denote the value at the $j$-th node in the $i$-th layer as $\omega_{ij}$. Then $\omega_{1j}$ is given by the $j$-th predictor input into the model, and for a $L$ layer neural network, $\omega_{L\cdot}$ is the predicted output. Take $\varphi_{i}(\cdot)$ to be the activation function in the $i$-th layer. Then for $i\in\{2,\dots,L\}$, $\omega_{ij} = \varphi_{i}\left(\sum_{\ell = 1}^{n^{[i-1]}}\nu_{\ell,j}^{[i-1]}\omega_{i-1,\ell}\right) = \varphi_{i}(a_{ij})$. In this expression, $\nu_{\ell,j}^{[i-1]}$ is the weight for the connection between the $\ell$-th node on layer $(i-1)$, and the $j$-th node on the $i$-th layer. A feed forward, fully connected neural network's architecture is thus characterized by the number of layers ($L$), the sizes of the layers ($n^{[i]}$), and the activation functions used at each layer ($\varphi_i(\cdot)$).

The activation functions throughout the hidden layers are selected to introduce nonlinearity into the model. While it is conceptually possible to use any function for this purpose, some see more use in practice. For instance, the \emph{ReLU} ($\varphi(x) = \max\{0,x\}$), the \emph{sigmoid} ($\varphi(x) = (1+\exp(-x))^{-1}$), or the \emph{tanh} ($\varphi(x) = \tanh(x)$) are all commonly used activation functions. In contrast to the hidden layers, the activation function for the output serves as a method for transforming the predicted values to the relevant scale or range of the output space. The activation function for the output layer can be thought of as analogous to the link function from a generalized regression framework. In fact, this similarity is more than aesthetic. If we take a two layer neural network, with $n^{[2]} = 1$ as the output layer, and define $\varphi_2(x) = g^{-1}(x)$, then the predicted outcome has the same form as in equation~(\ref{eq::outcome_model}). While the estimation procedure for regression coefficients differs from the procedure for estimating neural network weights, this demonstrates that neural networks are a strict generalisation of the introduced parametric regression framework.

 The process of \emph{fitting} a neural network is the process of estimating (or \emph{learning}) the sets of weights, $\nu_{\cdot}^{[\cdot]}$, based on a training data set, and a specified architecture. Typically, neural networks are fit via stochastic gradient descent (SGD) with backpropagation \cite{BP_SGD}. This is an efficient technique for optimizing a loss function over the space of network weights. SGD is a gradient descent technique which replaces the computed gradient across all of the data with an estimate of the gradient computed on a randomly selected sample from the data. This reduces the computational burden of updating throughout the optimization process, particularly when the gradients are extremely large. In practice, SGD replaces updates of the form $\nu' = \nu + \zeta\nabla\mathcal{L}$ with updates of the form \[\nu' = \nu + \frac{\zeta}{m}\sum_{i=1}^m\nabla\mathcal{L}_i,\] where $m$ is the size of the batches used for the approximation. Backpropagation is an algorithm which efficiently computes the gradient of the loss function with respect to the weights of the network.

In general, the larger the network, both in terms of the number of layers and the number of nodes within each layer, the more capable the neural network is of learning complex relationships. However, larger networks require more computational resources and more data to fit, and are more prone to overfitting than smaller networks \cite{Dropout}. As a result, selecting an architecture for a neural network involves searching for a sufficiently small neural network that can accurately capture the relationships which are present. A drawback to using neural networks is the cost of training. Neural networks generally require a large amount of data to fit in a stable manner, and can take an exceptionally long time, even on powerful hardware, to train until convergence (particularly when compared to, for instance, regression models with closed form solutions).

\subsection{Machine Learning Models in Dietary Research}\label{subsec::background}
There are growing efforts to leverage machine learning techniques in the analysis of dietary intake data. A wide array of techniques have been applied, demonstrating the potential utility of modern algorithms for understanding diet and health relationships, to date there has been less of an effort to ensure that these approaches are being applied appropriately with dietary intake data. While the consideration of measurement error for traditional statistical modeling for dietary intake data has received a great deal of attention, applications of machine learning to dietary intake data have largely ignored the effects of measurement error. This has been the case both for models that are predominantly concerned with prediction and for models which admit easier interpretations. In the machine learning literature, separate from nutrition research, concerns relating to measurement error have also been left unaddressed. 

Despite the underdeveloped theory, machine learning models have been investigated for dietary intake analysis. Morgenstern et al.\cite{Morgenstern_2022} use conditional inference forest models to relate reported nutrient intake on a given day to cardiovascular disease risk, building predictive models. Panaretos et al.\cite{Panaretos_2018} show that machine learning methods have better predictive performance, compared with linear regression, in the classification of long-term cardiometabolic risk when using individual's dietary patterns. Bodnar et al.\cite{Bodnar_2020} use the Super Learner algorithm to investigate the relationship between fruit and vegetable intake and adverse outcomes in pregnancy, and contrast these results with a more traditional multivariable logistic regression. Madeira et al.\cite{Madeira2020} use latent class transition models to assess the association between a posteriori derived dietary patterns and nutritional status in older adults. Similarly, Farmer et al.\cite{Farmer2020} use latent class analysis to explore dietary patterns emerging in US adults. Shang et al.\cite{Shang2020} use regression models, random forests, and gradient boosting machines to identify dietary determinants for changes to cadiometabolic risk in children. Wright et al.\cite{Wright2020} perform hierarchical clustering and dimension reduction through treelet transformations, to examine how dietary patterns are associated with periodontitis. Côté and Lamarche\cite{Review1} provide an overview of current and future applications of machine learning to dietary research, focusing on the ways that complex interactions and diverse data can be integrated using these techniques. 

The aforementioned research largely ignored the impacts of measurement error. Russo and Bonassi\cite{ReviewArticle} provide guidance to the application of neural networks in the context of nutritional epidemiology. With regards to the problem of measurement error, they discuss the use of technology to reduce errors that are present, emphasize that collecting high-quality data is key to the successful application of neural networks, and indicate that expanding the the size of data to increase statistical power may be useful in overcoming errors. We extend this advice to demonstrate that merely increasing the sample size will be insufficient to overcome the issues of measurement error. Morgenstern et al.\cite{Morgenstern_2021} discuss how advances in machine learning may be able to be used to overcome challenges faced in nutritional epidemiology, with a particular emphasis on problems of measurement error. Their focus is on how data collection can be improved through the application of machine learning techniques, minimizing the impact of errors from the outset. Collecting high quality data that are free from errors is preferable to trying to correct for the effects of measurement error, however, most dietary intake data that are being collected or have previously been collected will be subject to measurement error. As a result, techniques to ensure valid analyses in the presence of errors remain important.

\section{Considerations for Predictive modeling with Measurement Error}\label{sec::theory}
In this section, we investigate three separate theoretical considerations for the application of neural networks to error-prone data. In Section~\ref{subsec::additivity}, we consider data transformations which can be applied to error-prone variables to result in an approximately additive error structure. We demonstrate how, depending on the assumed measurement error model, these transformations can result in variance reduction when averaging error-prone predictors. Section~\ref{subsec::overfitting} addresses the impact that measurement error is likely to have on the capacity for a neural network to \emph{overfit} data. We argue that the problem of overfitting is likely to be exacerbated when the input data are subject to error. Finally, Section~\ref{subsec::sample_size} considers the how trading off an increased sample size for an increasing number of replicate measurements may result in an overall improvement to the predictive models performance. These results suggest that, contrary to existing advice in the literature, there are diminishing returns to an increased sample size when data are noisy, if sufficient numbers of replicate measurements have not been taken.

\subsection{Error Variance Reduction and Transformations to Additivity}\label{subsec::additivity}
When considering data measured with error, intuitively, the smaller the error variance, the better a predictive model will perform, as there is less noise distorting the relationship between the outcome and the informative components of the predictors. While it is not possible to directly reduce the error variance without changing the instrument being used for measurement, it is generally possible to simulate the effects of a variance reduction by averaging repeated measurements of the predictors. Suppose that we directly observe $X^*_{\ell,j}$ as in equation~(\ref{eq::surrogate_definition}), for $j=1,\dots,k$. The measurement error in this observation is captured through the term $\epsilon_{\ell,j}$ with variance $\sigma^2$. Take $f(x) = x$ and consider \[\overline{X^*}_{\ell}^{(+)} = \frac{1}{k}\sum_{j=1}^{k} X^*_{\ell,j} = \beta_{0,\ell} + \beta_{Z,\ell}'Z + u_{\ell} + \frac{1}{k}\sum_{j=1}^k \epsilon_{\ell,j} = \beta_{0,\ell} + \beta_{Z,\ell}'Z + u_{\ell} + \overline{\epsilon}_{\ell},\] then the measurement error of this variance term is given by the variance of $\overline{\epsilon}_\ell$, which is $\sigma^2/k$. As a result, when there is an additive structure for the measurement error, averaging provides a reduction in variance by a multiplicative factor of the number of days which are averaged. In other words, averaging is equivalent to using an improved instrument with a reduced variance. As alluded to in the introduction, as $k \to \infty$, the error term $\overline{\epsilon}_\ell$ vanishes.

Suppose instead that $f(x) = \log(x)$, so that the daily observed values are subject to an exponential transformation. Then \[\overline{X^*}_{\ell} = \frac{1}{k}\sum_{j=1}^{k} X^*_{\ell,j} = \exp\left\{\beta_{0,\ell} + \beta_{Z,\ell}'Z + u_{\ell}\right\}\left(\frac{1}{k}\sum_{j=1}^{k}\exp\left\{\epsilon_{\ell,j}\right\}\right).\] This results in a smaller variance reduction in the error term. Specifically, if we improved the instrument we were using to give the same variance as the average in this case, then this would correspond to an error term with variance \[\sigma_{\text{Reduced}}^2 = \log\left[\frac{1}{2}\left(\sqrt{\frac{k - 4e^{\sigma^2} + 4e^{2\sigma^2}}{k}}\right)+1\right].\] So long as $k \geq 2$, then we have that $\frac{\sigma^2}{k} \leq \sigma_\text{Reduced}^2 \leq \sigma^2$. Thus, the resulting error variance is reduced by averaging, however, the reduction in variance is smaller than the reduction for an additive model.

Suppose that instead of taking $\overline{X^*}_\ell$ under the exponential transform, we use a log-transform, and then re-transform the variables exponentially. That is, consider the quantity \[\widetilde{X^*}_\ell = \exp\left(\frac{1}{k}\sum_{j=1}^k \log(X^*_{\ell,j})\right) = \exp\left(\overline{X^*}_{\ell}^{(+)}\right) = \exp\left\{\beta_{0,\ell} + \beta_{Z,\ell}'Z + u_\ell\right\}\exp\left\{\overline{\epsilon}_\ell\right\}.\] From the above argument, we know that the error variance of this term is less than averaging on the un-transformed scale. In general, if the transformation $f$ is sufficiently smooth, $f(\overline{X^*}_\ell^{(+)})$ will have smaller variance in the measurement error term than $\overline{X^*}_\ell$.

\begin{lemma}\label{lemma::additivity}
 Subject to regularity conditions of $f$, then taking $\epsilon_i \stackrel{iid}{\sim} N(0,\sigma^2)$ independently of $\Omega$ for $i=1,\dots,k$, we have that \[\var\left[\left.f(\Omega + \overline{\epsilon})\right|\Omega\right] \leq \var\left[\left.\frac{1}{k}\sum_{i=1}^{k}f(\Omega+\epsilon_i)\right|\Omega\right].\]
\end{lemma}

Lemma~\ref{lemma::additivity} tells us that if the transformations we are considering are particularly well-behaved (for instance, exponential functions or polynomials), then it is the case that averaging over an additive model before re-applying the transformation will result in a quantity with less variance than averaging the observed quantities. The exact transformation $f$ will dictate the degree to which the variances are altered, however, it will generally be true that transformations to additivity reduce the variance of the underlying variates. Eckert et al.\cite{Eckert1997} discuss techniques for transforming error-prone variables to a scale on which they are additive, providing techniques for obtaining this variance reduction.

Whether this variance reduction is useful in predictive models depends on whether or not the $\epsilon_{\ell,j}$ are informative to the outcome of interest. In particular, if $E[Y|X^*, Z,u] = E[Y|Z,u]$, then the error terms are non-informative, and reducing their variance is useful for predictive accuracy. Given the specification that the true, long-term variates are $X_\ell = E[X^*_{\ell,j}|Z]$, this variance reduction may not prove useful for prediction. If instead we conceive of the relationship as being related to $X_\ell = f^{-1}(E[f(X^*_{\ell,j})|Z])$, then these notions are potentially useful in predictive modeling.

\subsection{Measurement Error and Neural Network Overfitting}\label{subsec::overfitting}
Overfitting refers to the problem of a predictive model performing substantially better on a training set compared with a validation set. For a particular estimation procedure resulting in the prediction model $\widehat{T}$, take the loss on the training set to be $\widetilde{\mathcal{L}}(\widehat{T}) = \widetilde{\mathcal{L}}$ and the loss on the testing set to be $\mathcal{L}(\widehat{T}) = \mathcal{L}$. Generally, we expect that $\widetilde{\mathcal{L}} < \mathcal{L}$, but in a well-calibrated model the difference should be minimal. As the size of the validation sample increases, equation~(\ref{eq::loss_function}) indicates that $\mathcal{L}$ will converge almost surely to $E[\mathcal{L}] = E\left[(Y - \widehat{T}(X^*, Z))^2\right]$. The degree of overfitting in a model can then be assessed by the degree to which $E[\mathcal{L}]$ exceeds $\widetilde{\mathcal{L}}$. 

The best predictor, in terms of MSE, for $Y$, observing $(X^*, Z)$ is taking $\widehat{T}(X^*, Z) = E[Y|X^*, Z]$. Doing so results in a minimal $E[\mathcal{L}]$ of $\var(Y|X^*, Z)$. When the true variates, $X$, are available, we get that $E[\mathcal{L}] \geq \sigma_Y^2$. Consider the outlined setting, without a Box-Cox transformation, so that $X^*_{\ell,j} = \beta_{0,\ell} + \beta_{Z,\ell}'Z + u_\ell + \epsilon_{\ell,j}$ and $X_\ell = \beta_{0,\ell} + \beta_{Z,\ell}'Z + u_\ell$. Here $X^*_{\ell,j} = X_\ell + \epsilon_{\ell,j}$. Considering the variance of $Y$ given $\{X^*, Z\}$ gives \[
 \var(Y|X^*, Z) = \var\left(\left.g^{-1}(\alpha_0 + \alpha_X'(X^* - \epsilon) + \alpha_Z'Z)\right|X^*, Z\right) + \sigma_Y^2 \geq 0 + \sigma_Y^2,
\] where the inequality follows since the first term has a residual random component which is not fully specified by $X^*$ and $Z$. This means that the lower bound on the validation loss when subject to measurement error, in this setting, will exceed the lower bound for the validation loss in the error-free setting.

Despite the fact that the theoretical limits of performance on testing data favour the error-free context, this says nothing of the capacity for neural networks to fit the training data. Previous work has shown that, for many architectures of neural networks, adding noise to the input data does not lead to an increased training loss \cite{Zhang2017}. In other words, even when there is no true pattern to learn, neural networks show a capacity to encode the training data exactly. In our context this means that, for a given neural network, we may not expect $\widetilde{\mathcal{L}}$ to differ substantially between the cases when $X$ are available or when $X^*$ are available. 

Combining these two results leads to the conclusion that, in the presence of measurement error, neural networks have a greater capacity to overfit the training data when compared to training on error-free data. This is a result that is confirmed in our simulation experiments (Section~\ref{sec::simulation_overfit}). Overfitting can be addressed through sufficient validation, regularization techniques, and carefully selected model architectures. However, the increased capacity to overfit presents challenges to the application of neural networks in dietary intake studies. Practitioners must be careful to ensure that techniques to prevent overfitting are applied, and that training performance does not lend undue confidence to the out-of-sample performance of the model. It is also worth emphasizing that overfitting does not mean that the performance of the neural network will be poor on out-of-sample data, or that it will suffer when compared to alternative techniques. Instead, overfitting renders the training error to be an incorrectly calibrated assessment of the true model performance, and this problem is exacerbated by the presence of measurement error.

\subsection{Sample Size and Replicate Counts for Predictive Power}\label{subsec::sample_size}
In order to increase the size of data that are used to train a predictive model, it is possible to increase either the sample size, $n$, or the replicate counts, $k$. Typically in dietary assessment research it will be the case that $n\gg k$. It is worth considering how these two quantities impact the performance of a predictive model. To do so, consider a simple case that is analytically tractable. Suppose that $Y = g^{-1}(X) + e$, with a univariate $X\sim N(\mu_X, \sigma_X^2)$, and $e\sim N(0,\sigma_Y^2)$ independently of all other random variables. Further, suppose that $X^* = X + V$ for $V \sim N(0,\sigma_V^2)$ independently of all other random variables. If $g^{-1}$ is sufficiently smooth in an interval around $E[X]$ such that $g^{-1}(X) \approx g^{-1}(E[X]) + \frac{d}{dX}g^{-1}(E[X])(X - E[X])$ from a first order Taylor series, then \[\var(g^{-1}(X)|X^*) \approx \left[\frac{d}{dX}g^{-1}(E[X])\right]^2\var(X|X^*).\] The best predictor for $Y$ given $X^*$ has a lower bounded loss of $\var(Y|X^*) = \var(g^{-1}(X)|X^*) + \sigma_Y^2$. By assumption, $X = X^* - V$, and as a result $\var(X|X^*) = \var(X^* - V|X^*) = \var(V|X^*)$. In this simplified setting we conclude that $X^* \sim N(\mu_X, \sigma_X^2 + \sigma_V^2)$ and that \[\var(V|X^*) = \frac{\sigma_V^2\sigma_X^2}{\sigma_V^2 + \sigma_X^2}.\]

Instead of $X^*$ being a single observation we take $X^* = \frac{1}{k}\sum_{j=1}^kX^*_j$, where each $X^*_j = X + V_j$, where each $V_j$ is independent and identically distributed according to $N(0,\sigma_V^2)$. Then $X^*$ will be normally distributed with mean $\mu_X$ and variance $\frac{1}{k}\sigma_V^2 + \sigma_X^2$, and \[\var(V|X^*) = \frac{\sigma_V^2\sigma_X^2}{\sigma_V^2 + k\sigma_X^2}.\] The difference from using repeated measurements comes in the form of a variance reduction, on the order of $1/k$. Put differently, as more replicates of $X^*_j$ are observed, the lower bound for the predictive loss continues to improve. As $k$ grows to $\infty$, $\var(Y|X^*)$ converges to $\var(Y|X)$, and prediction with the error-prone variates can be done as well as prediction with the error-free variates. This suggests that, in the event that the functional relationship $E[Y|X^*]$ is exactly known, it remains beneficial to increase the number of replicate observations for each individual in order to improve out-of-sample predictive performance.

Normally, the functional relationship defining $E[Y|X^*]$ will not be explicitly known, and will need to be estimated based on observed data. Suppose that $E[Y|X^*] = h(X^*)$, and that this relationship is consistently estimated from a sample of size $n$. Then \[E\left[(Y - \widehat{h}_n(X^*))^2|X^*\right] = E\left[(Y - h(X^*))^2|X^*\right] + E\left[(h(X^*) - \widehat{h}_n(X^*))^2|X^*\right].\] The first term is simply $\var(Y|X^*)$, exactly as in the previous analysis. The second term is the estimation MSE. While the structure of neural networks renders the task of finding bounds on convergence rates challenging, current theoretical findings in the deep learning literature suggest that, in at least some settings, deep learning can achieve near parametric optimal rates in terms of the MSE \cite{Rate1,Rate2,Rate3}. That is to say the second term, in the best case scenario, is of order $n^{-1 + \delta}$, for an arbitrarily small $\delta > 0$. 

Combining these two results, the residual prediction MSE, in an idealized scenario, will be $O\left(k^{-1} + n^{-1+\delta}\right)$. That is, \[E\left[(Y - \widehat{h}_n(X^*))^2|X^*\right] = \sigma_Y^2 + \frac{C_1}{k} + \frac{C_2}{n^{1-\delta}},\] for constants $C_1$ and $C_2$. In dietary research, we have a comparatively small $k$, often with $k\leq 2$, and a comparatively large $n$, often with $n$ in the thousands. Correspondingly, even when $C_2 \gg C_1$, the scaling factors are likely to be such that an increase in $k$ will have a more sizable decrease in the overall error than an increase in $n$. In fact, when these results hold, it may be the case that halving $n$ and doubling $k$ such that $n\times k$ remains constant may provide a more efficient allocation of resources in terms of overall predictive performance. This is a result which we explore in more depth in simulation experiments (Section~\ref{sec::simulation_n_k_tradeoff}). The general pattern we observe, as indicated by this heuristic theory, is that when a large enough sample is taken for adequate model performance, it is advantageous for model performance to increase $k$, even if that means reducing $n$. 

Two important considerations stem from these results. First, it is critical to understand the convergence behaviour of neural networks in the presence of measurement error. It is not sufficient to take a large $n$ to overcome the impacts of measurement error on prediction. Even in theoretically infinite samples, the models that are fit to data that are subject to measurement error will perform worse than models fit to data that are free from error. Depending on the exact setting, this degradation in performance can be substantial. Instead, convergence to the optimal model requires the simultaneous increase of both $n$ and $k$. As a result, when studies are to be designed, it remains important to sufficiently budget for repeated measurements, even when prediction is the goal. Generally, dietary intake data with $k\leq 2$, owing to cost and data quality constraints. These results suggest that, even for predictive modeling, study designs should consider achieving higher levels of replication whenever possible. Second, when a model is fit with a set number of replicates, $k$, future predictions implicitly assume the same level of replication. If a predictive model is fit using data from $2$ days of replicates, it will not be advantageous to record more than $2$ days of information for out-of-sample prediction. Additionally, prediction will rely on being able to collect $2$ days of replicated information out-of-sample as well. This requirement can be overcome through distributional assumptions and transformations, however, in practice it is prudent to design the study with these considerations in mind.

\section{Measurement Error in Neural Networks}\label{sec::numerical_studies}
We now consider numerical experiments to investigate the impacts of measurement error on the predictive performance of neural networks applied to data that are subject to error. We begin by illustrating how the potential benefits of using neural networks for prediction may be undermined in the presence of measurement error. We then give evidence that increasing the number of available replicate measurements can lead to improved predictive performance. Finally, we show how, if there is sufficiently large measurement error, neural networks do not outperform incorrectly specified linear regression models, even in highly nonlinear settings. Taken together, these results suggest that, while there is a strong motivation for finding valid methods of applying neural networks to the analysis of dietary intake data, analysts must proceed with caution in the application of these techniques.

\subsection{Transformations and Overfitting}\label{sec::simulation_overfit}
The first scenario assumes that the only variates are two error-prone variates, $X_1$ and $X_2$. We take $h(X)=\frac{X_1}{X_2}$ so that \[Y = \alpha_0 + \alpha_1\frac{X_1}{X_2} + e,\] where $e \sim N(0, \sigma_Y^2)$, independent of all other variates. We assume that we observe two days' worth of replicates, giving for each individual $\{X^*_{11}, X^*_{12}, X^*_{21}, X^*_{22}\}$, where we generate \[X^*_{\ell,j} = \beta_{0,\ell} + u_{\ell} + \epsilon_{\ell,j},\] where $\begin{bmatrix}u_1 & u_2\end{bmatrix}'\sim N(0, \Sigma_u)$ and $\begin{bmatrix}\epsilon_{1,j} & \epsilon_{2,j}\end{bmatrix}' \sim N(0, \Sigma_\epsilon)$ independent of all variates for each $j$, and we take \[X_\ell = E\left\{\left[\lambda X^*_{\ell,j}+1\right]^{1/\lambda}\right\}. \] We take $\sigma_Y^2 = 1$, $\beta_{01} = 36$, $\beta_{02} = 27.5$, $\alpha_0 = 98.5$, $\alpha_1 = 4.0$, $\Sigma_U = \begin{bmatrix}20 & 15.5 \\ 15.5 & 25.5\end{bmatrix}$, $\Sigma_\epsilon = \begin{bmatrix}38 & 20.5 \\ 20.5 & 34.5\end{bmatrix}$, and $\lambda = 0.35$. We consider replicating measurements for $2, 4, 6, 8$, and $10$ days, with a sample size of $12000$ (giving $12000$ observations per day).

\begin{figure}
 \centering
 \includegraphics[width=\textwidth]{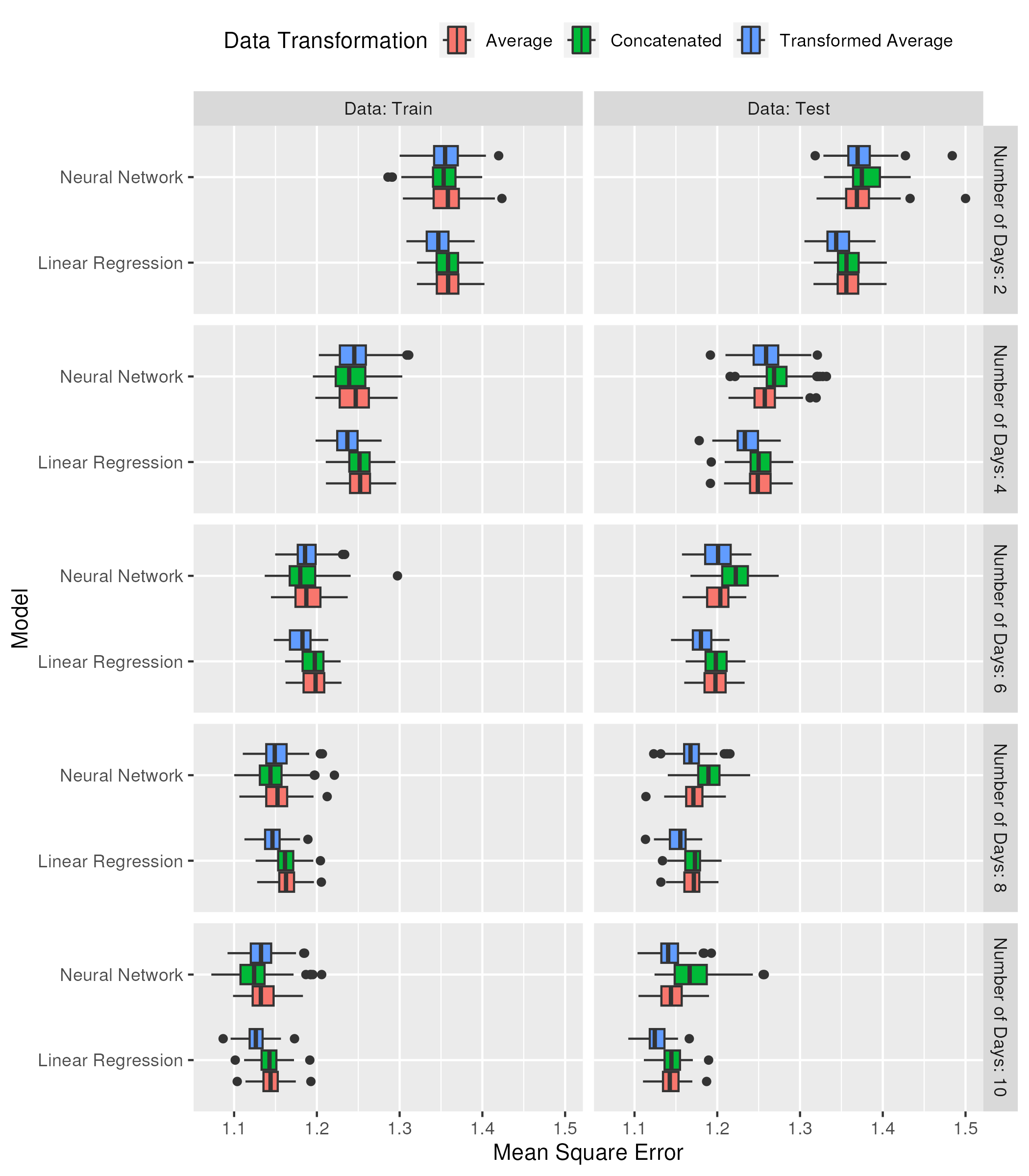}
 \caption{MSE for both neural networks and linear regression models predicting an outcome based on the ratio of predictors, where each predictor is subject to additive measurement error, and the sample size is $12000$. Results are shown for both the training and testing data, and over an increasing number of days, where each individual has measurements taken for the predictors each day.}
 \label{fig::simulation_1}
\end{figure}

The analysis proceeds comparing linear regression models with neural networks. For each method we consider three different data transformations to prepare the data: \begin{enumerate}[label=(\arabic*)]
 \item averaging the error prone measurements across each day of observation (averaging);
 \item treating each observed, error prone measurement as a separate predictor (concatenation); or
 \item performing a Box-Cox transformation of the error prone measurements, and then averaging these results (transformed averaging).
\end{enumerate} Note, the linear regression model is incorrectly specified in that $X^*_1$ and $X^*_2$ are passed as linear predictors, rather than as a ratio. The neural network also takes each of the predictors linearly. Figure~\ref{fig::simulation_1} contains the MSE for each of the various scenarios, plotted by number of days, model type, and data preparation. The results are included for both the training data and for the testing data.

The results demonstrate the expected pattern that, as more data are used to predict outcomes, the overall MSE decreases both in the training and testing sets. The neural networks tended to perform better than the corresponding linear regression model in the training data, and performed worse than the linear regression models on the testing data, though these differences tend to be fairly minor. The neural network seems to overfit the training data, no matter the number of days, when each of the predictors are provided independently (concatenation), as evidenced by the fact that the concatenated results are the strongest in the training set and the weakest in the testing set. This is inline with the theoretical observations, where in the training data the neural network is able to use noise as a signal, despite the fact that this does not generalize to the testing data. With the linear regression, using the transformed average consistently outperformed all other techniques in terms of MSE. Within the neural networks, both averaging approaches performed similarly, and these tended to outperform the concatenated results on the test set.

\subsection{Sample Size and Replicate Counts}\label{sec::simulation_n_k_tradeoff}
\begin{figure}
 \centering
 \includegraphics[width=\textwidth]{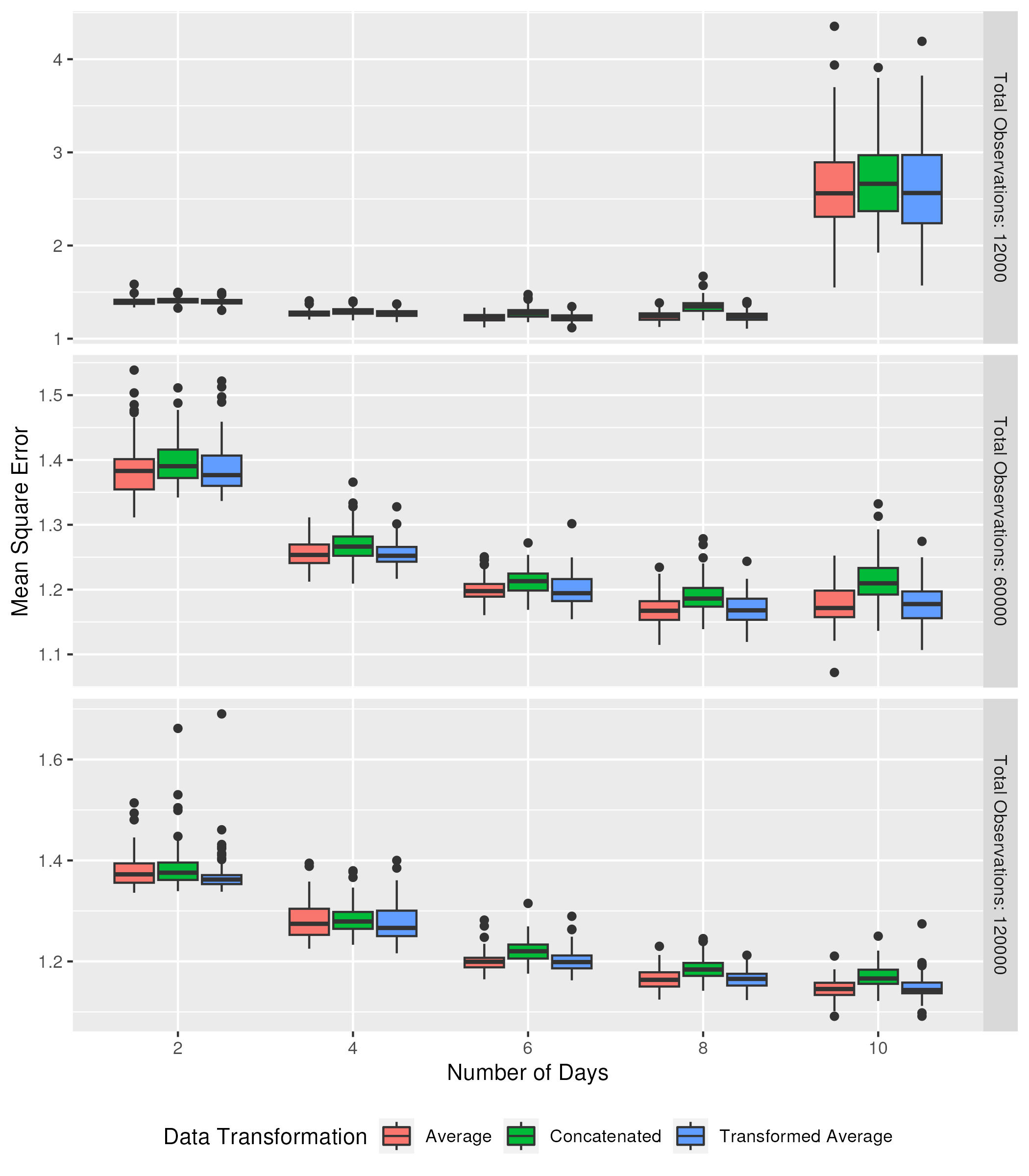}
 \caption{MSE for neural networks predicting an outcome based on the ratio of predictors, where each predictor is subject to additive measurement error. The total number of observations is fixed to $12000$, $60000$, or $120000$, and divided evenly between differing number of days. Results are shown for the testing data.}
 \label{fig::simulation_2}
\end{figure}

In the second simulation experiment, we consider the same scenario as in the previous subsection, taking $\lambda = 0.5$ rather than $\lambda = 0.35$, increasing the skewness of the simulated data. In these simulations, we vary the number of replicated days that we are averaging over, keeping the total number of replicates available across the sample. This allows us to investigate the trade-off between the benefits of a larger sample with the benefits of smaller error variances. In particular, we consider using $2,4,6,8$, or $10$ days' worth of replicates. We fix the total number of measurements to be either $12000$ (with samples sizes $6000, 3000, 2000, 1500$, or $1200$), $60000$ (with samples sizes $30000, 15000, 10000, 7500$, or $6000$), or $120000$ (with sample sizes $60000, 30000, 20000, 15000$, or $12000$). In Figure~\ref{fig::simulation_2} the MSE for the neural network, with each of the data transformations outlined in simulation 1, is shown. 

The results indicate that, all else equal, it is preferable to increase the number of days of replicates of the error-prone variables, even if that means sacrificing the total sample size. This pattern is evident when using $60000$ or $120000$ total observations, but not when using $12000$, suggesting that there is a minimum sample size required for the neural network to provide stable prediction. This suggests that it is worth considering the minimal sample sizes required to achieve stable results for specific scenarios, prior to deploying these methodologies. While not shown, regression analyses were run, with results similar to those presented in the first simulation. Additionally, the same overfitting of the concatenated results occurs. Comparing $60000$ and $120000$ total observations, the results are fairly similar, indicating that there may be a point of diminishing returns for this model.

\subsection{Prediction of Outcomes with Non-linearity}
In the third simulation experiment, we consider a scenario which is depends on nonlinear functions of the truth, for the error-prone predictors. The outcome is given by \[Y = \alpha_0 + \alpha_1X_1 + \alpha_2X_2 + \alpha_3\frac{X_1}{X_2} + \alpha_4\log(X_1) + \alpha_5\log(X_2) + \alpha_6\sqrt{X_1X_2} + Z_3 + e,\] where $e\sim N(0,1)$ independently of all other factors. We take $Z$ to be multivariate normal, with mean $0$ and covariance \[\Sigma_Z = \begin{bmatrix}
 1.28 & 0.21 & -0.07 & 0.32 \\
 0.21 & 1.98 & 1.28 & 0.34 \\
 -0.07 & 1.28 & 1.91 & 1.22 \\
 0.32 & 0.34 & 1.22 & 1.20
\end{bmatrix}.\] We generate $X = \begin{bmatrix}
 X_1 & X_2
\end{bmatrix}'$ to be drawn from the conditional distribution \[X|Z \sim N\left(\beta\begin{bmatrix} \mathbf{1} & 
 Z_1 & Z_2 & Z_3 & Z_4
\end{bmatrix}', \begin{bmatrix}
 20 & 15.5 \\ 15.5 & 25.5
\end{bmatrix}\right).\] We take the error-prone observations to be $X^*_{\ell,j} = X_{\ell} + \epsilon_{\ell,j}$, for $j=1,2$, and $\begin{bmatrix}
 \epsilon_{1,j} & \epsilon_{2,j}
\end{bmatrix}' \sim N(0,\Sigma_\epsilon)$ independently of all other variables, where $\Sigma_\epsilon = \begin{bmatrix}
 38 & 20.5 \\ 20.5 & 34.5
\end{bmatrix}$. We take a sample size of $n=40000$ for the training and testing samples, and repeat the simulations $100$ times. 

\begin{figure}
 \centering
 \includegraphics[width=\textwidth]{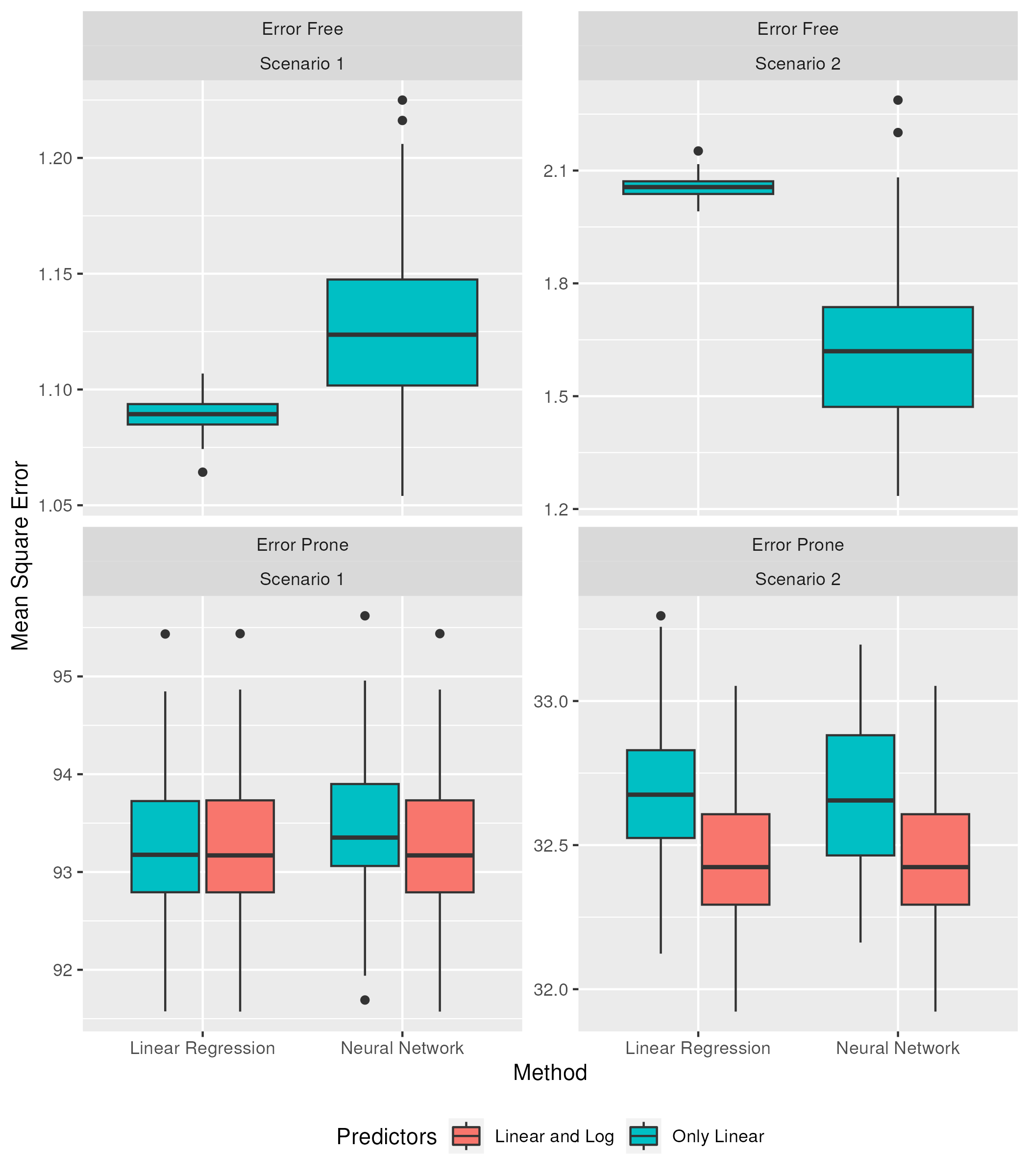}
 \caption{MSE for linear regression and neural networks predicting an outcome based on a nonlinear function of predictors, for both the error free and error prone settings. Two separate scenarios are considered, each using two days of replicates. Results are shown for the testing data.}
 \label{fig::simulation_3}
\end{figure}

Two scenarios are formed, varying $\alpha$ and $\beta$. Scenario 1 takes $\alpha = \begin{bmatrix}350 & 2 & -1 & 3 & 2 & 1 & -4\end{bmatrix}'$, and $\beta = \begin{bmatrix}
 100 & 2 & 0 & -1 & 0.5 \\
 100 & 0 & 2 & 1 & -0.5
\end{bmatrix}$. Scenario 2 takes $\alpha = \begin{bmatrix}350 & 1 & -1 & 50 & 25 & 25 & -1\end{bmatrix}'$, and $\beta = \begin{bmatrix}
 50 & 2 & 0 & -1 & 0.5 \\
 50 & 0 & 2 & 1 & -0.5
\end{bmatrix}$. For each scenario we consider both error-free and error-prone models. For the error-free models, we consider only the true predictors as linear terms. For the error-prone models, we consider both the models which include the error-prone predictors as linear terms, and ones which also include the log transformed predictors. These predictors are averaged over the two days of observations. The MSE for each technique across both scenarios on the test data is displayed in Figure~\ref{fig::simulation_3}.

Across both error-prone scenarios the linear regression and neural network perform comparably when using the same data predictors. For both methods, inclusion of the log terms improves the MSE. This difference is predictably more pronounced in scenario 2 where the log terms have a larger impact on the outcome. These results are best understood when compared to the error-free context. Both the linear regression and neural network are able to attain near the theoretical lower bound on MSE in scenario 1, when using only the linear predictors. For scenario 2, the neural network substantially outperforms the linear regression, when considering only linear predictors. This improvement is entirely degraded, however, with the introduction of measurement error.

\section{Data Analysis}
To demonstrate the caution required when applying deep learning predictive models to noisy data, we consider an analysis of a subset of NHANES data from the 2005-2010 waves \cite{NHANES}. These data are collected using 24HRs, with processes that have adapted and evolved over time based recommendations and emerging data needs \cite{Ahluwalia2016}. Generally, two 24HRs are attempted to be collected for each individual, however, some respondents have missing second measurements.

The considered subset is inspired by the analysis conducted by Zhang et al.\cite{Zhang2013}. These authors investigated the association between usual sodium and potassium intakes and blood pressure. Adults aged $20$ and older with reliable information on their 24HRs (where reliability is determined by the data collectors) were included in the data ($N=15702$). Individuals who were pregnant, missing blood pressure measures, missing the self-reported hypertension diagnosis indicator, reported being on a low sodium diet, or reported taking anti-hypertensive medication were excluded ($N=5134$ were excluded). The self-reported intakes of sodium and potassium (measured in mg) were included as the error-prone predictors, with up to two measurements taken per individual. The average diastolic blood pressure measurements were taken as the outcome. The predictors that are assumed to be measured without appreciable error are the individual's reported gender, age, cardiovascular disease status, educational attainment (beyond twelve years or less than twelve years total), reported ethnicity, reported smoking status, reported diabetes status, and an indicator for reported kidney disease status. Individuals with any missing, error-free predictors were excluded from the analysis ($N=14$ further individuals were excluded). In total, there were $N=10554$ individuals in the final data.

Once collected, the data were artificially split into a training set ($N=8442$) and a testing set ($N=2112$). These data were partitioned to include equivalent proportions of individuals who had a single repeated measure of each sodium and potassium ($N=1076$ and $N=270$ in the training and testing sets, respectively). Note, any individual who had a second measure for sodium had one for potassium, and vice versa. Once split, the training data were used to fit and select linear regression models and neural networks. These fitted models were evaluated on the testing set in terms of the MSE.

The neural networks used had two layers, the learning rate was set to $0.001$, and early stopping criteria were used. Prior to fitting, the data were standardized (to have $0$ mean and unit variance). The missing values (for the second day observations, no other covariates were missing) were imputed as by taking the first day's observations. Neural networks were fit using both the averaging and concatenation techniques discussed in the paper, with no further data transformations applied. For cross validation, $10\%$ of the data was used to validate. These analyses were performed in Python, using the PyTorch library \cite{PyTorch}. 

Three separate linear regression models were fit to the data. The first two took the data as they had been modified for the neural network (one using the concatenation, one using the average of the two days) and fit a simple linear model, with linear terms and no interactions. The third linear regression model followed a more rigorous modeling procedure, as would likely be applied in practice. First, a multivariate power transform was fit to the bivariate predictors, resulting in approximately normally distributed variates. These transformed responses were averaged for all individuals, and a predictor containing the number of observations (either $1$ or $2$) was included. Then, starting from a model that contained all identifiable three-way interactions, a backwards search procedure occurs to select the best subset model based on $10$-fold cross validation. This procedure is implemented with the R packages leaps\cite{leaps} and caret\cite{caret}. Running this procedure selected a model that contained $44$ terms in addition to the intercept, where, in addition to predictors themselves, two- and three-way interactions are considered a single term. Sensitivity of these results were consider by taking the smallest model that was within $0.01$ of the $44$ term model in terms of cross validated root mean square error (the square root of MSE). This lead us to consider a $19$ term model as well. 

The resulting MSE for each of the models on both the training and testing data are shown in Table~\ref{tab::nhanes_analysis}. First, we note that the performance of the linear regression model selected through the backwards selection procedure which kept only $19$ terms was nearly identical to that of the backwards selection procedure which was optimal and kept $44$ terms, demonstrating a lack of sensitivity. Moreover, we can see that both of the naive regression models performed worse (on the training and testing data) than the more rigorously justified regression model, as is anticipated from the theory. 

Comparing the neural networks and regression models, the neural networks (using either averaged or concatenated data) performed slightly worse than the regression models on the testing set, despite improved training performance. A stark difference exists for the concatenated results which outperformed all other techniques on the training data, and was less effective than either of the selected regression or the neural network with averaged data on the testing set. This is indicative of the same overfitting behaviour demonstrated in the numerical studies. It is important to note that this overfitting behaviour was observed even when taking explicit steps to avoid its impact.

\begin{table}
 \centering
 \begin{tabular}{lcc}
 Model & Training MSE & Testing MSE \\ \hline
 Neural Network (Averaged) & $130.1669$ & $142.5655$\\
 Neural Network (Concatenated) & $\mathbf{129.0694}$ & $142.8862$\\
 Linear Regression (Averaged) & $138.7071$ & $146.8728$\\
 Linear Regression (Concatenated) & $138.7010$ & $146.7932$\\
 Linear Regression (Backwards Selection, Size=$44$) & $132.2561$ & $\mathbf{142.0518}$\\
 Linear Regression (Backwards Selection, Size=$19$) & $132.4637$ & $\mathbf{142.0518}$ \\ \hline 
 \end{tabular}
 \caption{The training and testing MSE for the neural networks and linear regression models fit to the NHANES data, based on an $80\%$/$20\%$ train/test split. Bolded values indicate the most performant techniques for the training and testing sets, respectively.}
 \label{tab::nhanes_analysis}
\end{table}

While performance is comparable between the neural networks and linear regression models, it is important to consider the costs of applying machine learning in this setting. The neural networks require substantially more computational time and resources to fit, and produce results which are less interpretable than the linear regression. Moreover, the fitted neural network models do not lend themselves to inferential procedures, meaning that prediction is the only utility of the model. Given that the performance is not improved in this setting, and in fact, seems to be slightly degraded, caution in the application of these techniques is warranted.

\section{Discussion}\label{sec::discussion}
Dietary intake data are complex and error-prone. There is an ongoing effort to leverage the powerful techniques emerging from machine learning research to conduct analysis of diet and health relationships, and this research has shown promising results. However, concerns specifically relevant to dietary intake data, particularly as they relate to measurement error, have not been previously addressed in the statistical or computing literature. The most prominent modern machine learning techniques, including artificial neural networks, are primarily predictive models. The literature on the impact of measurement error on prediction models is sparse, owing largely to the fact that in many settings it can be simply be ignored. However, the errors that are common in dietary intake data are often large enough to dramatically reduce the performance of prediction models, resulting in a strong motivation to consider avenues for improvement. This is salient when these predictive models are intended to be used to make decisions or predict health outcomes of individuals and populations using dietary intake data. 

We systematically investigate the impact of classical measurement error on the performance of neural networks in settings that are analogous to those confronted in nutrition epidemiology research. Even though this preliminary work does not attempt to address the complex measurement error structures comprising systematic bias as well as noise that are thought to apply to self-report dietary data, we demonstrate that even well-behaved measurement error is cause for concern. We show that, as expected, error reduces model performance for neural networks, and this reduction can be sizable. We demonstrate how, when these errors are sufficiently large, the signal becomes nearly undetectable, rendering misspecified linear regression models equally effective at out-of-sample prediction. We illustrate theoretically and via empirical simulations that neural networks are prone to overfitting data that are subject to measurement error, and avoiding this requires particular care during model development.

Beyond the shortcomings of neural networks, however, we also saw the promise of improved performance at detecting complex relationships. To this end, we showed that models can benefit dramatically from increasing replicate data. In particular, we showed that for predictive models, there are likely to be diminishing returns in increasing the sample size beyond a particular point; however, increasing the number of measurements per individual can continue to improve the model performance. When designing a study that will be used to fit a predictive model, it is important to consider the number of replicates that are able to be measured, even if this results in a decreased overall sample size. In practice, increasing the number of observations per individual results in reduced data quality, which may undermine the observed results. Our results stand in contrast to standard practice in the nutrition surveillance literature which suggest that two recalls suffices for data collection. Instead, these results lend further credibility to authors who have suggested that more than two replicates are preferred, even accounting for the degradation of data quality\cite{Schatzkin2003}. Further research that considers how to make use of data which distributional shift, or that takes further takes into account alternative dietary assessment mechanisms (such as food frequency questionnaires) may help to resolve this theory-practice gap. Using FFQs alongside 24HRs presents alternative concerns relating to the systematic biases present in FFQs. Still, this combined approach has been investigated with some promise for traditional techniques\cite{Freedman2018}. We illustrated how, depending on the underlying structure of the data, transformations to additivity may serve to improve the variance reduction attained through averaging procedures. These transformations have been well-studied in the measurement error literature, and they may serve an important role in making the most of the available replicates.

Overall, this work serves to stress the important point that, while neural networks are powerful semiparametric prediction models, they are subject to the same concerns present in traditional statistical methodologies. Not only that, but these techniques tend to be computationally expensive, and nearly uninterpretable. When they are able to deliver superior predictive performance, these tradeoffs are likely worthwhile. However, our work demonstrates that, without care to the structure of the data that are available, it is unlikely that theoretically defensible results can be reliably obtained in dietary research by applying machine learning techniques directly. Our work serves to demonstrate some preliminary areas of focus for the valid application of machine learning techniques to data that are prone to measurement error, and leaves open many questions regarding the best approach to leveraging the power of neural networks for nutritional epidemiology research. Further simulation and theoretical work is required to expand these results, and in particular, to make concrete recommendations for overcoming the outlined issues. In the interim, however, care is required when deploying black box prediction models in these settings.

\bibliographystyle{unsrtnat}
\bibliography{references} 

\clearpage

\appendix 

\setcounter{page}{1}

\section{Conditions on \texorpdfstring{$f$}{f}: Proof of Lemma 1}
Suppose that there is a convergent Taylor series representation of $f(x)$ such that \[f(x) = \sum_{r=1}^\infty \frac{f^{(r)}(a)}{r!}\left(x - a\right)^r.\] We wish to compare the conditional variance of $f(\Omega + \overline{\epsilon})$ with $\overline{f(\Omega + \epsilon)}$, given $\Omega$. The evident choice is to take $a = \Omega$, so that \begin{align*}
 f(\Omega + \overline{\epsilon}) &= \sum_{r=1}^\infty \frac{f^{(r)}(\Omega)}{r!}\overline{\epsilon}^r \\
 \overline{f(\Omega + \epsilon)} &= \frac{1}{k}\sum_{j=1}^k\sum_{r=1}^\infty \frac{f^{(r)}(\Omega)}{r!}\epsilon_j^r \\
 \var&\left[\left.f(\Omega + \overline{\epsilon})\right|\Omega\right] \\
 &= \sum_{r=1}^\infty\sum_{r'=1}^\infty \cov\left(\left.\frac{f^{(r)}(\Omega)}{r!}\overline{\epsilon}^r,\frac{f^{(r')}(\Omega)}{r'!}\overline{\epsilon}^{r'}\right|\Omega\right) \\
 &= \sum_{r=1}^\infty\sum_{r'=1}^\infty \frac{f^{(r)}(\Omega)f^{(r')}(\Omega)}{r!(r')!} \cov\left(\overline{\epsilon}^r,\overline{\epsilon}^{r'}\right) \\
 &= \sum_{r=1}^\infty\sum_{r'=1}^\infty \frac{f^{(r)}(\Omega)f^{(r')}(\Omega)}{r!(r')!} \left(E[\overline{\epsilon}^{r+r'}]-E[\overline{\epsilon}^r]E[\overline{\epsilon}^{r'}]\right) \\
 &= \sum_{r=1}^\infty\sum_{r'=1}^\infty \frac{f^{(r)}(\Omega)f^{(r')}(\Omega)}{r!(r')!}\left(\frac{\sigma}{\sqrt{k}}\right)^{r+r'}\left(I(r+r'\text{ even})(r+r')!! - I(r,r'\text{ even})r!!(r')!!\right) \\
 &= \sum_{r=1}^\infty\sum_{r'=1}^\infty C(\Omega, r, r')\frac{\sigma^{r+r'}}{k^{(r+r')/2}} \\
 \var&\left[\left.\overline{f(\Omega + \epsilon)}\right|\Omega\right] \\ 
 &= \frac{1}{k^2}\sum_{j=1}^k\sum_{r=1}^\infty\sum_{j'=1}^k\sum_{r'=1}^\infty \cov\left(\left.\frac{f^{(r)}(\Omega)}{r!}\epsilon_j^r,\frac{f^{(r')}(\Omega)}{(r')!}\epsilon_{j'}^{r'}\right|\Omega\right) \\
 &= \frac{1}{k^2}\sum_{j=1}^k\sum_{r=1}^\infty\sum_{r'=1}^\infty \cov\left(\left.\frac{f^{(r)}(\Omega)}{r!}\epsilon_j^r,\frac{f^{(r')}(\Omega)}{(r')!}\epsilon_{j}^{r'}\right|\Omega\right) \\
 &= \frac{1}{k}\sum_{r=1}^\infty\sum_{r'=1}^\infty \cov\left(\left.\frac{f^{(r)}(\Omega)}{r!}\epsilon^r,\frac{f^{(r')}(\Omega)}{(r')!}\epsilon^{r'}\right|\Omega\right) \\
 &= \sum_{r=1}^\infty\sum_{r'=1}^\infty \frac{f^{(r)}(\Omega)f^{(r')}(\Omega)}{r!(r')!}\frac{\sigma^{r+r'}}{k}\left(I(r+r'\text{ even})(r+r')!! - I(r,r'\text{ even})r!!(r')!!\right) \\
 &= \sum_{r=1}^\infty\sum_{r'=1}^\infty C(\Omega, r, r')\frac{\sigma^{r+r'}}{k}.
\end{align*} Consider that, for $k \geq 2$, and $k > 1$, $k^k \geq k$, and so $\frac{1}{k^k} \leq \frac{1}{k}$. Noting that $r+r' \geq 2$ for all $r,r'$ in the summation, we get that \[C(\Omega,r,r')\frac{\sigma^{r+r'}}{k^{(r+r')/2}} \leq C(\Omega,r,r')\frac{\sigma^{r+r'}}{k} \implies \var\left[\left.f(\Omega + \overline{\epsilon})\right|\Omega\right] \leq \var\left[\left.\overline{f(\Omega + \epsilon)}\right|\Omega\right].\]

\end{document}